\documentclass[twocolumn]{article}
\usepackage[utf8]{inputenc}
\usepackage{graphicx}
\usepackage{amsmath}
\usepackage{hyperref}
\usepackage{natbib}
\usepackage{fancyhdr}
\usepackage{geometry}
\usepackage{titlesec}
\usepackage{lipsum}
\usepackage{orcidlink} 
\usepackage{bm}  
\usepackage{amsmath}
\usepackage{amssymb}
\setcitestyle{authoryear}  
\usepackage{booktabs}
\usepackage{placeins}
\hypersetup{
	colorlinks=true,   
	linkcolor=blue,   
	filecolor=blue,   
	citecolor=blue,   
	urlcolor=black     
}

\fancypagestyle{plain}{ 
	\fancyhf{}
}

\title{\vspace{1em}\hrule height 1.5pt \vspace{0.8em}Half-VAE: An Encoder-Free VAE to Bypass Explicit Inverse Mapping\vspace{0.8em}\hrule height 1.5pt\vspace{1em}}
\author{
	\begin{minipage}{.3\textwidth}
		\centering
		\small 
		\textbf{Yuan-Hao Wei\orcidlink{0000-0001-9439-0780}}\\
		Hong Kong Polytechnic University \\
		\texttt{Yuan-Hao.Wei@outlook.com}
	\end{minipage}
	\hfill
	\begin{minipage}{.3\textwidth}
		\centering
		\small 
		\textbf{Yan-Jie Sun\orcidlink{0000-0002-7967-6382}}\\
		Hong Kong Polytechnic University \\
		\texttt{yanjie.sun@connect.polyu.hk}
	\end{minipage}
	\hfill
	\begin{minipage}{.3\textwidth}
		\centering
		\small 
		\textbf{Chen Jason Zhang\orcidlink{0000-0002-3306-9317}}\\
		Hong Kong Polytechnic University \\
		\texttt{jason-c.zhang@polyu.edu.hk}
	\end{minipage}
}
\date{} 

\begin{document}
	\maketitle
	\thispagestyle{plain} 
	
	\begin{abstract}
	Inference and inverse problems are closely related concepts, both fundamentally involving the deduction of unknown causes or parameters from observed data. Bayesian inference, a powerful class of methods, is often employed to solve a variety of problems, including those related to causal inference. Variational inference, a subset of Bayesian inference, is primarily used to efficiently approximate complex posterior distributions. Variational Autoencoders (VAEs), which combine variational inference with deep learning, have become widely applied across various domains. This study explores the potential of VAEs for solving inverse problems, such as Independent Component Analysis (ICA), without relying on an explicit inverse mapping process. Unlike other VAE-based ICA methods, this approach discards the encoder in the VAE architecture, directly setting the latent variables as trainable parameters. In other words, the latent variables are no longer outputs of the encoder but are instead optimized directly through the objective function to converge to appropriate values. We find that, with a suitable prior setup, the latent variables, represented by trainable parameters, can exhibit mutually independent properties as the parameters converge, all without the need for an encoding process. This approach, referred to as the Half-VAE, bypasses the inverse mapping process by eliminating the encoder. This study demonstrates the feasibility of using the Half-VAE to solve ICA without the need for an explicit inverse mapping process.
	\end{abstract}
	
	\section{Introduction}
	Inverse problems typically involve deducing the system's input or parameters (underlying causes) from observed data (known information). An inverse problem refers to the process of inferring a system's input, parameters, or structure from observed outcomes. The directly observed information, such as images, sounds, and changes in the surrounding environment, are usually the result of multiple factors acting together (\cite{hyvarinen2019nonlinear}; \cite{pearl2019seven}; \cite{khemakhem2020variational}; \cite{locatello2020weakly}; \cite{scholkopf2021toward}; \cite{yang2021causalvae}; \cite{lachapelle2022disentanglement}; \cite{scholkopf2022causality}; \cite{lippe2022citris}). When variational inference is applied to inverse problems, it transforms the problem into one of probabilistic inference. Variational Autoencoders (VAEs) combine the principles of variational inference with deep learning (\cite{kingma2013auto}; \cite{rezende2014stochastic}; \cite{kingma2019introduction}), endowing them with strong nonlinear mapping capabilities, which in turn provide the potential to tackle complex inference problems. Although causal reasoning, disentanglement, and Independent Component Analysis (ICA) often correspond to different mathematical tools, VAEs have shown remarkable adaptability across these domains (\cite{higgins2017beta}; \cite{burgess2018understanding}; \cite{chen2018isolating}; \cite{kim2018disentangling}; \cite{casale2018gaussian}; \cite{ramchandran2021longitudinal}; \cite{brehmer2022weakly}; \cite{tonekaboni2022decoupling}; \cite{ahuja2023interventional}; \cite{wei2024innovative}; \cite{wendong2024causal}). This demonstrates the powerful generalization capability of variational inference as a mathematical tool when augmented by neural networks. Additionally, obtaining independent latent variables through variational inference can enhance the interpretability of the model, thereby endowing the generative process of VAEs with stronger logic and controllability as a generative model.
	
	Due to the integration of variational Bayesian theory, the encoding process in VAE can be regarded as an inference process. When VAE is designed to solve problems such as ICA or disentanglement, this inference process can be equivalent to an inverse mapping. Briefly, when considering the latent variable \( \mathbf{Z} \) as independent components, the encoding process in a VAE represents the inverse mapping from the mixed observations to the independent components, i.e., \( \mathbf{Z} = f^{-1}(\mathbf{X}) \). On the other hand, the decoding process in a VAE represents the remapping of the independent components back into the observed data, i.e., \( \mathbf{X} = f(\mathbf{Z}) \). Generally, \( f \) represents static mixing, which means the mixing process remains constant and does not vary over time or space. In many previous studies, methods using Variational Autoencoders (VAEs) for performing disentanglement or Independent Component Analysis (ICA) typically have a prerequisite that the dimensionality \( M \) of the input data \( \mathbf{X} \) is at least as large as the dimensionality \( N \) of the latent variable \( \mathbf{Z} \) \cite{casale2018gaussian, wei2024innovative}. This requirement, where \( M \geq N \), is crucial because the encoder struggles to handle underdetermined situations, where \( M \) is smaller than \( N \). In such underdetermined cases, the theoretical inverse mapping \( {f}^{-1} \) does not exist \cite{comon1994independent, cardoso1998blind, bofill2001underdetermined, hyvarinen2001independent, cichocki2002adaptive}.Therefore, a strategy is needed to bypass the explicit inverse mapping process in the VAE architecture.
	
	There has been some research exploring tasks by removing the encoder from the encoder-decoder architecture. \cite{bond2020gradient} investigated the generalization capability and reconstruction quality of VAEs after removing the encoder. \cite{shi2020unsupervised} designed an encoder-free network to address 3D shape description and retrieval, primarily using Maximum Likelihood Estimation for optimizing neural networks instead of variational inference. \cite{cervantes2022implicit} proposed Implicit Neural Representations to generate variable-length human motion sequences, employing an encoder-free architecture as well. These studies demonstrate the viability of encoder-free architectures in certain domains. However, no research has yet explored using an encoder-free architecture to bypass the explicit inverse process in solving ICA problems. Considering that the inverse mapping \( {f}^{-1} \), represented by the encoding process, does not exist under underdetermined conditions, one possible approach is to circumvent this challenging mapping by omitting the encoder. 
		
	This study discards the encoder in the VAE architecture, allowing the posterior distributions of the latent variables to be directly obtained through trainable parameters converging based on the objective function. This encoder-free VAE is referred to as Half-VAE. We use several randomly generated independent and identically distributed (i.i.d.) signals as the independent components and select Gaussian Mixture Models (GMM) (\cite{dempster1977maximum}; \cite{mclachlan2000finite}; \cite{bishop2006pattern}; \cite{reynolds2009gaussian}) as the prior distributions, with the GMM parameters set as optimizable, adaptive parameters. In this design, the parameters of the posterior and prior distributions of the latent variables, as well as the decoder's parameters, are all optimized through the objective function. From the perspective of parameter optimization, a typical VAE requires optimizing the parameters of both the encoder and decoder, whereas Half-VAE directly optimizes the posterior distributions of the latent variables. We find that even without an encoder, the different dimensions of the Half-VAE's latent variables can still converge to mutually independent i.i.d. sequences. Discarding the encoder is equivalent to discarding the explicit inverse mapping stage, leaving only the 'forward' process from the independent components to the observed signals.  
	
	The primary contribution of this study is the introduction of the Half-VAE, an encoder-free VAE architecture designed to avoid explicit inverse mapping in solving the ICA problem. This research derives the variational lower bound of the Half-VAE and uses it as the objective function. Through simulation experiments, the effectiveness of the Half-VAE in addressing the ICA problem has been validated, laying a foundation for future research.

	\section{Half-VAE}
	
		\subsection{Derivation}
		For the derivation of the objective function of Half-VAE, the first step is to maximize the log-marginal likelihood of the observed data \( \mathbf{X} \):
		\begin{equation}
			\ln p_{\boldsymbol{\theta}}(\mathbf{X}) = \ln \int p_{\boldsymbol{\theta}}(\mathbf{X}, \mathbf{Z}) \, d\mathbf{Z},
		\end{equation}
		where \( p_{\boldsymbol{\theta}}(\mathbf{X}, \mathbf{Z}) \) is the joint distribution of the observed data \( \mathbf{X} \) and the latent variables \( \mathbf{Z} \). 	
		Introducing an implicit distribution  \(I(\mathbf{Z})\) to rewrite the log marginal likelihood:
		\begin{equation}
			\ln p_{\boldsymbol{\theta}}(\mathbf{X}) = \ln \int \frac{p_{\boldsymbol{\theta}}(\mathbf{X}, \mathbf{Z})}{I(\mathbf{Z})} I(\mathbf{Z}) \, d\mathbf{Z},
		\end{equation}
		where \(I(\mathbf{Z})\) represent the latent variable distributions characterized by trainable parameters.
		Since the \( \ln \) function is concave, we can use Jensen's Inequality (\cite{jensen1906fonctions}; \cite{cover1999elements}; \cite{boyd2004convex}):
		\begin{equation}
			\ln p_{\boldsymbol{\theta}}(\mathbf{X}) \geq \int I(\mathbf{Z}) \ln \frac{p_{\boldsymbol{\theta}}(\mathbf{X}, \mathbf{Z})}{I(\mathbf{Z})} \, d\mathbf{Z}.
		\end{equation}
		This inequality represents the evidence  lower bound (ELBO) of the log-marginal likelihood.
		We can further expand the joint distribution \( p_{\boldsymbol{\theta}}(\mathbf{X}, \mathbf{Z}) \) as a product of the conditional distribution \( p_{\boldsymbol{\theta}}(\mathbf{X}|\mathbf{Z}) \) and the prior distribution \( p(\mathbf{Z}) \):
		\begin{equation}
			\ln p_{\boldsymbol{\theta}}(\mathbf{X}) \geq \int I(\mathbf{Z}) \ln \frac{p_{\boldsymbol{\theta}}(\mathbf{X}|\mathbf{Z}) p(\mathbf{Z})}{I(\mathbf{Z})} \, d\mathbf{Z}.
		\end{equation}
		Breaking down the above equation into two terms:
		\begin{equation}
			\mathbb{E}_{I(\mathbf{Z})}[\ln p_{\boldsymbol{\theta}}(\mathbf{X}|\mathbf{Z})] = \int I(\mathbf{Z}) \ln p_{\boldsymbol{\theta}}(\mathbf{X}|\mathbf{Z}) \, d\mathbf{Z},
		\end{equation}
		and
		\begin{equation}
			- D_{\text{KL}}(I(\mathbf{Z}) \| p(\mathbf{Z})) = \int I(\mathbf{Z}) \ln \frac{I(\mathbf{Z})}{p(\mathbf{Z})} \, d\mathbf{Z},
		\end{equation}
		where the first term represents the reconstruction error, i.e., the expected log-likelihood of generating the data \( \mathbf{X} \) given the latent variables \( \mathbf{Z} \); the second term is the Kullback-Leibler (KL) divergence between the implicit posterior distribution \( I(\mathbf{Z}) \) and the prior distribution \( p(\mathbf{Z}) \), which measures how much \( I(\mathbf{Z}) \) deviates from \( p(\mathbf{Z}) \). Finally, the full expression for the variational lower bound can be obtained:
		\begin{equation}
			\ln p_{\boldsymbol{\theta}}(\mathbf{X}) \geq \mathbb{E}_{I(\mathbf{Z})}[\ln p_{\boldsymbol{\theta}}(\mathbf{X}|\mathbf{Z})] - D_{\text{KL}}(I(\mathbf{Z}) \| p(\mathbf{Z})).
		\end{equation}
		The above ELBO equation represents a lower bound that need maximizing by optimizing the parameters \( \boldsymbol{\theta} \) and \( I(\mathbf{Z}) \). The goal is for \( I(\mathbf{Z}) \) to approximate the true posterior distribution as closely as possible, while ensuring that the log-likelihood of generating the observed data \( \mathbf{X} \) is maximized.
		
		This derivation demonstrates that, even without an explicit encoding process, we can still obtain a variational lower bound similar to that of a standard VAE (\cite{kingma2013auto}) by introducing an implicit distribution \( I(\mathbf{Z}) \) determined with trainable parameters. This lower bound is used for optimizing the Half-VAE.
		
		In fact, in the VAE loss function, maximizing the ELBO forces \( D_{\text{KL}}(q(\mathbf{Z}|\mathbf{X}) \| p(\mathbf{Z}|\mathbf{X})) \) to approach zero, thereby making \( q(\mathbf{Z}|\mathbf{X}) \) approximate \( p(\mathbf{Z}|\mathbf{X}) \). Here, \( p(\mathbf{Z}|\mathbf{X}) \) represents the true posterior of the latent variables, while \( q(\mathbf{Z}|\mathbf{X}) \) is an additional term introduced for approximation. In typical VAEs, this introduced \( q(\mathbf{Z}|\mathbf{X}) \) is represented by the encoding process, which maps directly from \( \mathbf{X} \) to \( \mathbf{Z} \). However, in the theoretical derivation of variational inference, the term used to approximate \( p(\mathbf{Z}|\mathbf{X}) \) does not necessarily have to be \( q(\mathbf{Z}|\mathbf{X}) \). In this study, \( I(\mathbf{Z}) \) is directly used to approximate \( p(\mathbf{Z}|\mathbf{X}) \). Although the explicit encoding process is removed, the core principle of variational inference remains unchanged.
		
		\subsection{Architecture}
		Assume that \( \mathbf{X} \in \mathbb{R}^{M \times L} \) is a observable sample with \( M \) features, where \( L \) denotes the length of the features across the different dimensions. We aim to decompose this observable sample into mutually independent components \( \mathbf{Z} \in \mathbb{R}^{N \times L} \), where \( N \) represents the number of independent components. Within the framework of variational Bayesian inference, the independent components \( \mathbf{Z} \in \mathbb{R}^{N \times L} \) are expressed probabilistically, with means \( \mathbf{Z}_\mu \in \mathbb{R}^{N \times L} \) and variances \( \mathbf{Z}_\sigma \in \mathbb{R}^{N \times 1} \). Figures 1 and 2 illustrate the schematic diagrams of the process for solving the ICA problem using the traditional VAE architecture and the Half-VAE architecture, respectively. In a VAE, the mean and variance of the latent variables are produced by the encoder, with the encoding process functioning as an explicit inverse mapping. In contrast, in the Half-VAE, the latent variables are treated as trainable parameters that are directly optimized based on the objective function. While the goal of the Half-VAE is also to infer the distribution of the latent variables, it achieves this without an explicit inverse mapping process.
		\begin{figure}[ht]
			\centering
			\includegraphics[width=0.5\textwidth]{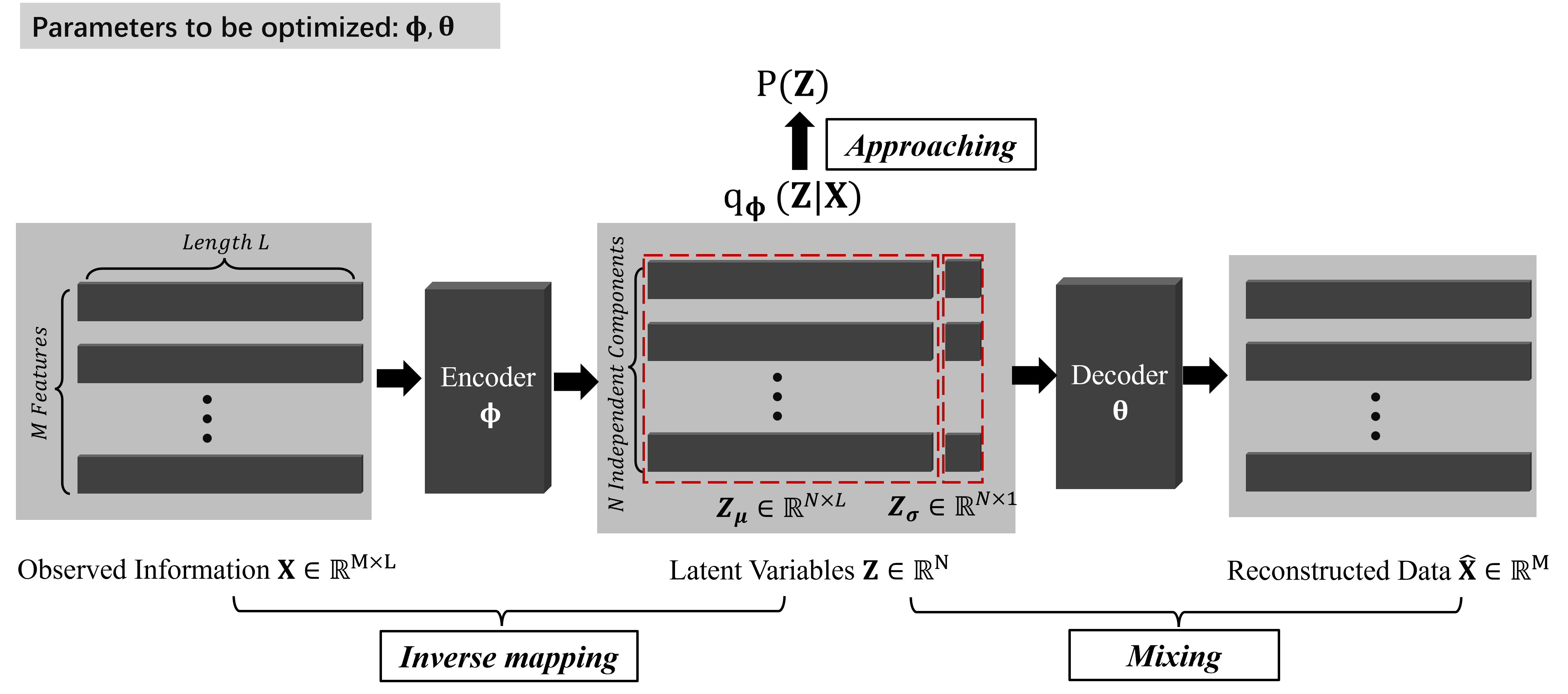}
			\caption{Schematic of VAE}
			\label{Schematic of VAE}
		\end{figure}
		\begin{figure}[ht]
			\centering
			\includegraphics[width=0.5\textwidth]{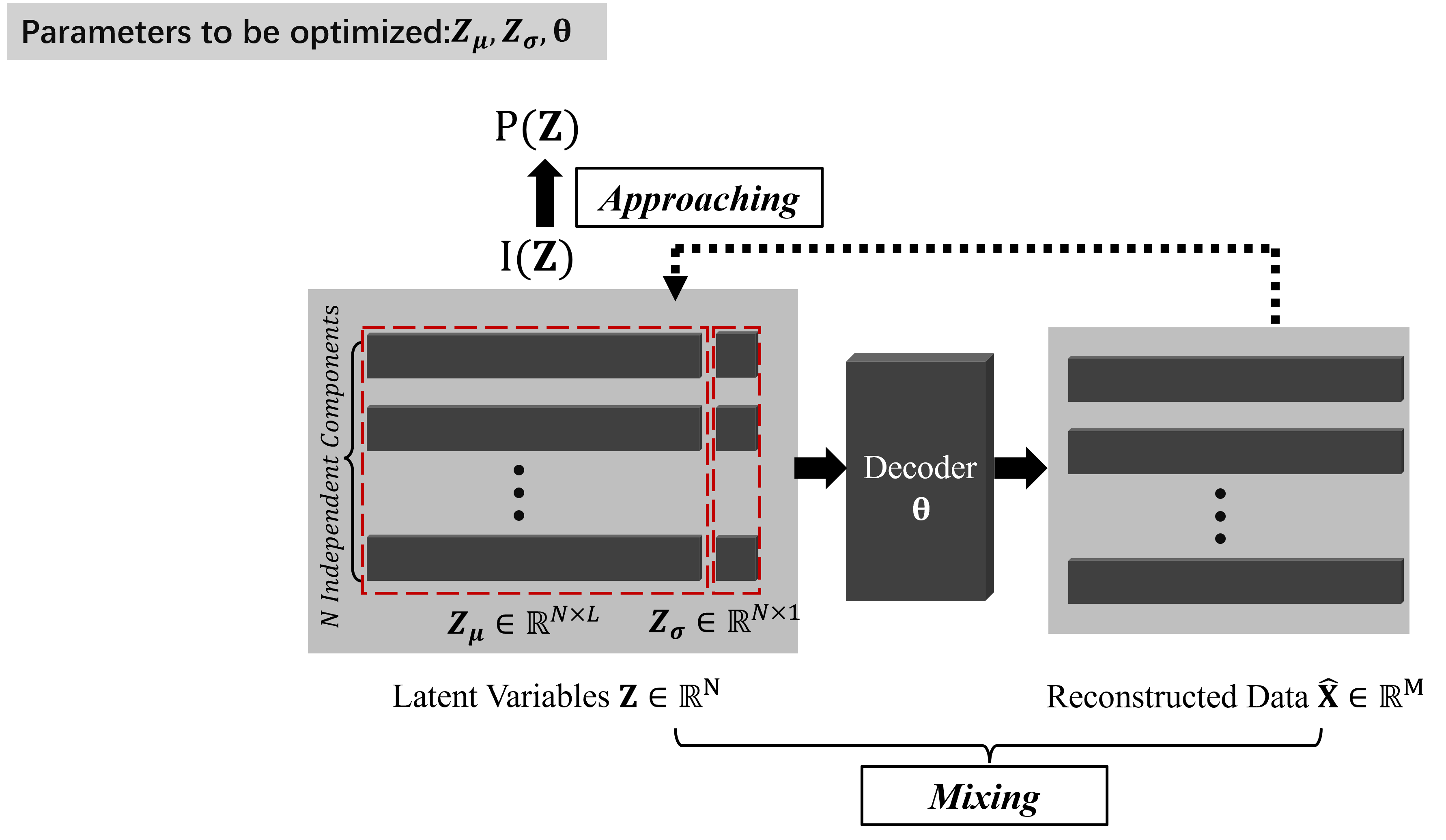}
			\caption{Schematicof Half-VAE}
			\label{Schematicof Half-VAE}
		\end{figure}	
		
		\subsection{GMM prior}
		From the objective function (Equation 7), it becomes clear that a well-suited prior distribution is crucial for effective inference of the latent variables $\bm{Z}$. Therefore, instead of using the standard normal distribution commonly employed in typical VAE framework (\cite{kingma2013auto}), we opt for GMMs as the priors for the latent variables. The standard normal distribution lacks the flexibility needed to capture the complex, unknown distributions of the independent components. In contrast, GMMs, as powerful probabilistic models, excel at fitting complex distributions and are not restricted by the limitations of a single Gaussian component, making them more adaptable to intricate data structures and flexible distributional assumptions.
		
		A Gaussian Mixture Model (GMM) is defined as:
		\begin{equation}
			\begin{split}
				p(\mathbf{Z}) = \sum_{k=1}^{K} \pi_k \mathcal{N}(\mathbf{Z}|m_k, \Sigma_k)
			\end{split}
		\end{equation}
		where \( \pi_k \) are the mixture weights, \( \mu_k \) are the means, and \( \Sigma_k \) are the covariances for each component \( k \) of the GMM. Given that the true distributions of the independent components in the ICA problem are unknown, we cannot directly assign precise parameters to the GMM priors. However, we can design these parameters \( \boldsymbol{\pi} \), \( \boldsymbol{m} \), and \( \boldsymbol{\Sigma} \) as trainable variables, allowing their values to be optimized through the objective function. For simplicity, we set the weights \( \boldsymbol{\pi} \), means \( \boldsymbol{m} \), and covariances \( \boldsymbol{\Sigma} \) are elements of the parameter set \( \boldsymbol{\Psi} \), i.e., \( \boldsymbol{\pi}, \boldsymbol{m}, \boldsymbol{\Sigma} \in \boldsymbol{\Psi} \).
		Consequently, the loss functions for the VAE and Half-VAE with GMM priors can be further expressed as follows:		
		\begin{equation}
			\begin{split}
				L_{\text{VAE}} (\boldsymbol{\phi}, \boldsymbol{\theta}, \boldsymbol{\Psi}; \mathbf{X}) = 
				- \mathbb{E}_{q_{\boldsymbol{\phi}} (\mathbf{Z}|\mathbf{X})} \left[ \ln p_{\boldsymbol{\theta}} (\mathbf{X}|\mathbf{Z}) \right] \\
				+ \lambda \cdot D_{\text{KL}} \left[ q_{\boldsymbol{\phi}} (\mathbf{Z}|\mathbf{X}) \| \text{GMM}_{\boldsymbol{\Psi}} (\mathbf{Z}) \right],
			\end{split}
		\end{equation}
		\begin{equation}
			\begin{split}
				L_{\text{Half-VAE}} (\mathbf{Z}_\mu, \mathbf{Z}_\sigma, \boldsymbol{\theta}, \boldsymbol{\Psi}; \mathbf{X}) = 
				- \mathbb{E}_{I(\mathbf{Z})} \left[ \ln p_{\boldsymbol{\theta}} (\mathbf{X}|\mathbf{Z}) \right] \\
				+ \lambda \cdot D_{\text{KL}} \left( I(\mathbf{Z}) \| \text{GMM}_{\boldsymbol{\Psi}} (\mathbf{Z}) \right),
			\end{split}
		\end{equation}
		where \( \lambda \) is used to adjust the weights of each item in the loss functions. The \( I(\mathbf{Z}) \) directly represented by \( \mathbf{Z}_\mu\) and \( \mathbf{Z}_\sigma\). Note that, for the convenience of performing stochastic gradient descent in neural networks, we have reformulated the maximization of the ELBO (see Equation 7) into the minimization of the negative ELBO as loss functions, as shown in Equations 9 and 10. It is important to emphasize that each independent component \( \mathbf{Z}_i \) has its corresponding prior \( \text{GMM}_{\boldsymbol{\Psi}_i} \). Since the components \( \mathbf{Z}_i \) are mutually independent, the total KL divergence term in Equations (9) and (10) can be further expressed as the sum of the KL divergences for each independent component:
		\begin{equation}
			\begin{split}
				D_{\text{KL}} \left[ q_{\boldsymbol{\phi}} (\mathbf{Z}|\mathbf{X}) \| \text{GMM}_{\boldsymbol{\Psi}} (\mathbf{Z}) \right] = \\
				\sum_{i=1}^{N} D_{\text{KL}} \left( q_{\boldsymbol{\phi}} (\mathbf{Z}_i|\mathbf{X}) \| \text{GMM}_{\boldsymbol{\Psi}_i} (\mathbf{Z}_i) \right),
			\end{split}
		\end{equation}
		\begin{equation}
			\begin{split}
				D_{\text{KL}} \left( I(\mathbf{Z}) \| \text{GMM}_{\boldsymbol{\Psi}} (\mathbf{Z}) \right) = 
				\sum_{i=1}^{N} D_{\text{KL}} \left( I(\mathbf{Z}_i) \| \text{GMM}_{\boldsymbol{\Psi}_i} (\mathbf{Z}_i) \right).
			\end{split}
		\end{equation}
		It is worth noting that the GMM is just one of many available prior distributions. For the Half-VAE, different types of priors can be considered when dealing with signals of varying characteristics.

	\section{Experiments}
	
	To evaluate the ability of the proposed Half-VAE to solve the ICA problem, we design a set of simulation experiments. First, multiple random signals are generated, as shown in Figure 3. During the generation process, efforts are made to make these signals as independent from each other as possible. After passing through a mixing mapping, we obtain the mixed observations, as illustrated in Figure 4. Without knowledge of the ground truth, the independent components are inferred in an unsupervised manner using both the VAE and Half-VAE.
	\begin{figure}[htbp]
		\centering
		\includegraphics[width=0.5\textwidth]{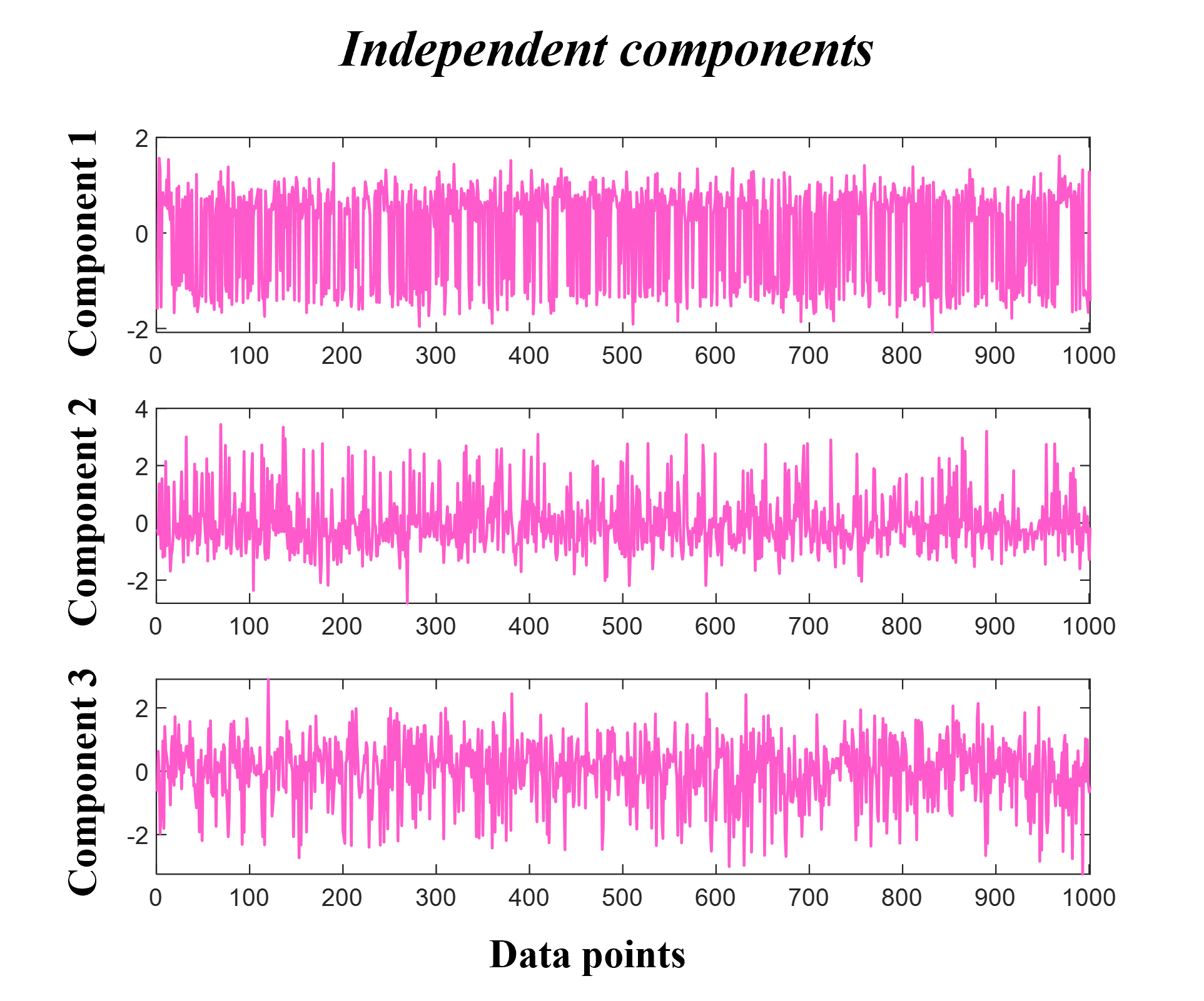}
		\caption{Independent components}
		\label{Independent components}
	\end{figure}
	\begin{figure}[h]
		\centering
		\includegraphics[width=0.5\textwidth]{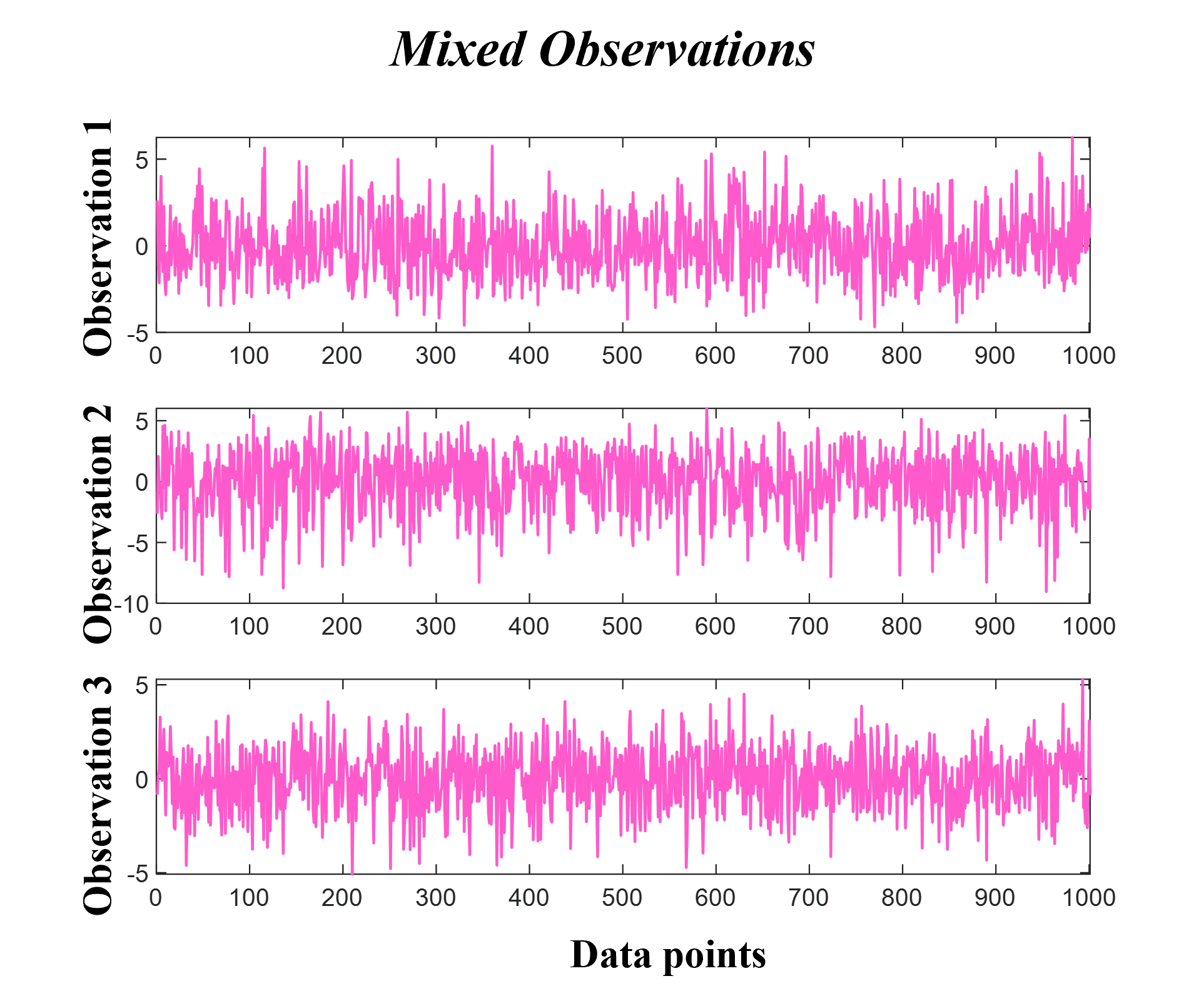}
		\caption{Observations}
		\label{Observations}
	\end{figure}
	
	The number of Gaussian distributions in the GMM prior for each independent component is empirically set to 3, i.e., \( k=3 \). The parameters of both the VAE and Half-VAE are optimized according to their loss functions , respectively. Figure 5 illustrates the evolution of \( \mathbf{Z}_\mu \) over epochs. It can be observed that \( \mathbf{Z}_\mu \) gradually approaches the true independent components as the Half-VAE model is optimized effectively.
	\begin{figure}[htbp]
		\centering
		\includegraphics[width=0.5\textwidth]{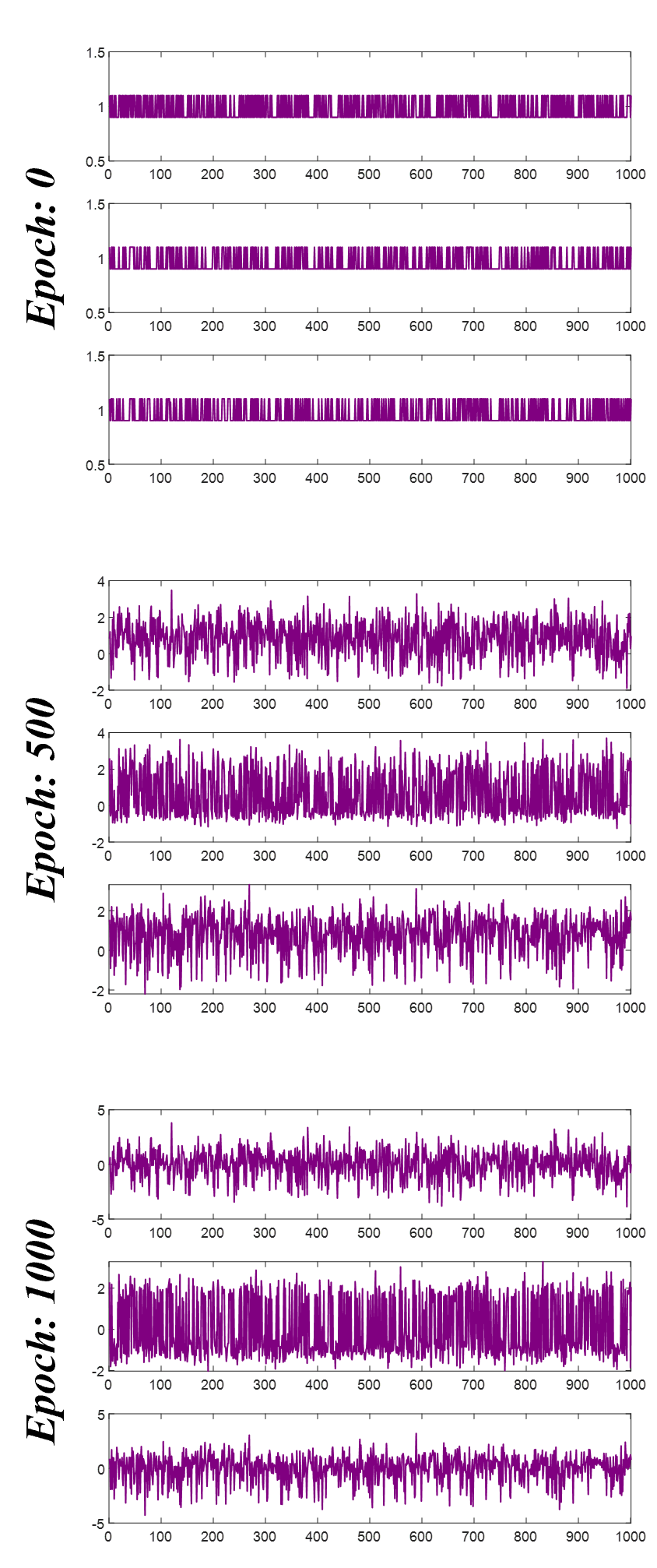}
		\caption{The change of \(\mathbf{Z}_\mu\) with epoch}
		\label{figzmuepoch}  
	\end{figure}
	
	It is important to note that in both Half-VAE and VAE, the latent variables are represented probabilistically, meaning that the inferred independent components are distributions rather than fixed values. Figure 6 shows the distribution of each independent component inferred by the Half-VAE after the loss function has converged, denoted as \( I(\mathbf{Z}) \). The pink lines represent the mean of each independent component, while the gray shaded areas indicate the 95\% confidence intervals. It can be observed that the model exhibits relatively low uncertainty in the estimated results, as the confidence intervals are quite narrow.
	\begin{figure}[htbp]
		\centering
		\includegraphics[width=0.5\textwidth]{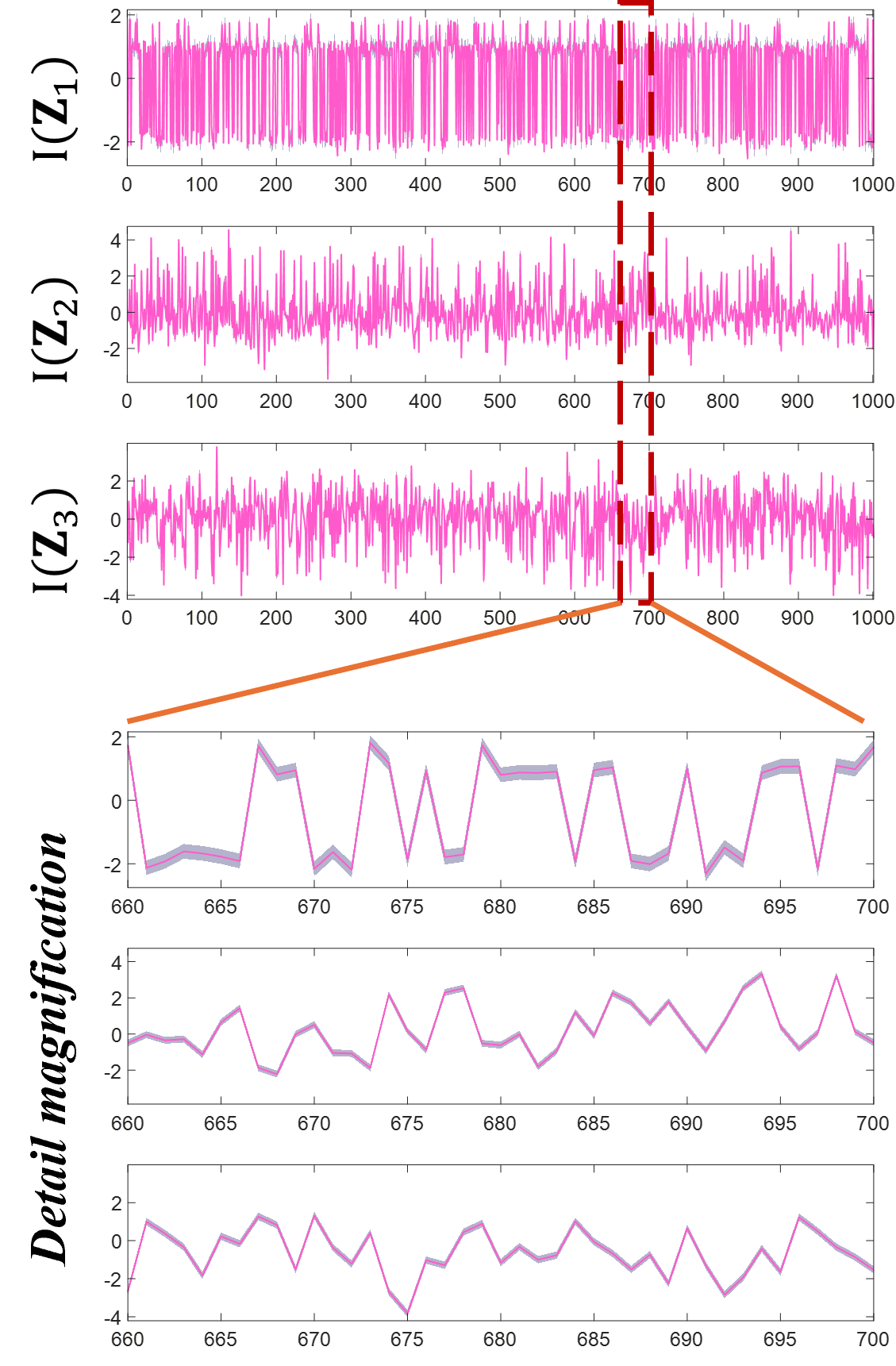}
		\caption{Estimated distributions of independent components}
		\label{Estimated distributions of independent components}
	\end{figure}
	
	The inference results of the Half-VAE and VAE are shown in Figures 7 and 8. For a broader comparison, we also evaluate the performance of a vanilla VAE, following the original settings for ICA. The prior of the vanilla VAE is set to a standard normal distribution (\cite{kingma2013auto}) rather than a GMM.
	The results for the vanilla VAE are shown in Figure 9. Due to the inherent scaling ambiguity in ICA solutions (\cite{comon1994independent}; \cite{hyvarinen2001independent}), we apply the z-score normalization method (\cite{altman1992introduction}; \cite{iglewicz1993volume}) to both the true independent components and the estimated results to compare them on the same scale. In these figures, the purple solid lines represent the normalized ground truth independent components, while the green dashed lines represent the means of the inferred independent component distributions for each model. It can be observed that both the Half-VAE and the VAE with GMM priors accurately infer the independent components, whereas the results from the vanilla VAE deviate significantly from the ground truth. The root mean square errors (RMSE) between the inferred results and the corresponding ground truth for each model are presented in Table 1.
	
	\begin{figure}[htbp]
		\centering
		\includegraphics[width=0.5\textwidth]{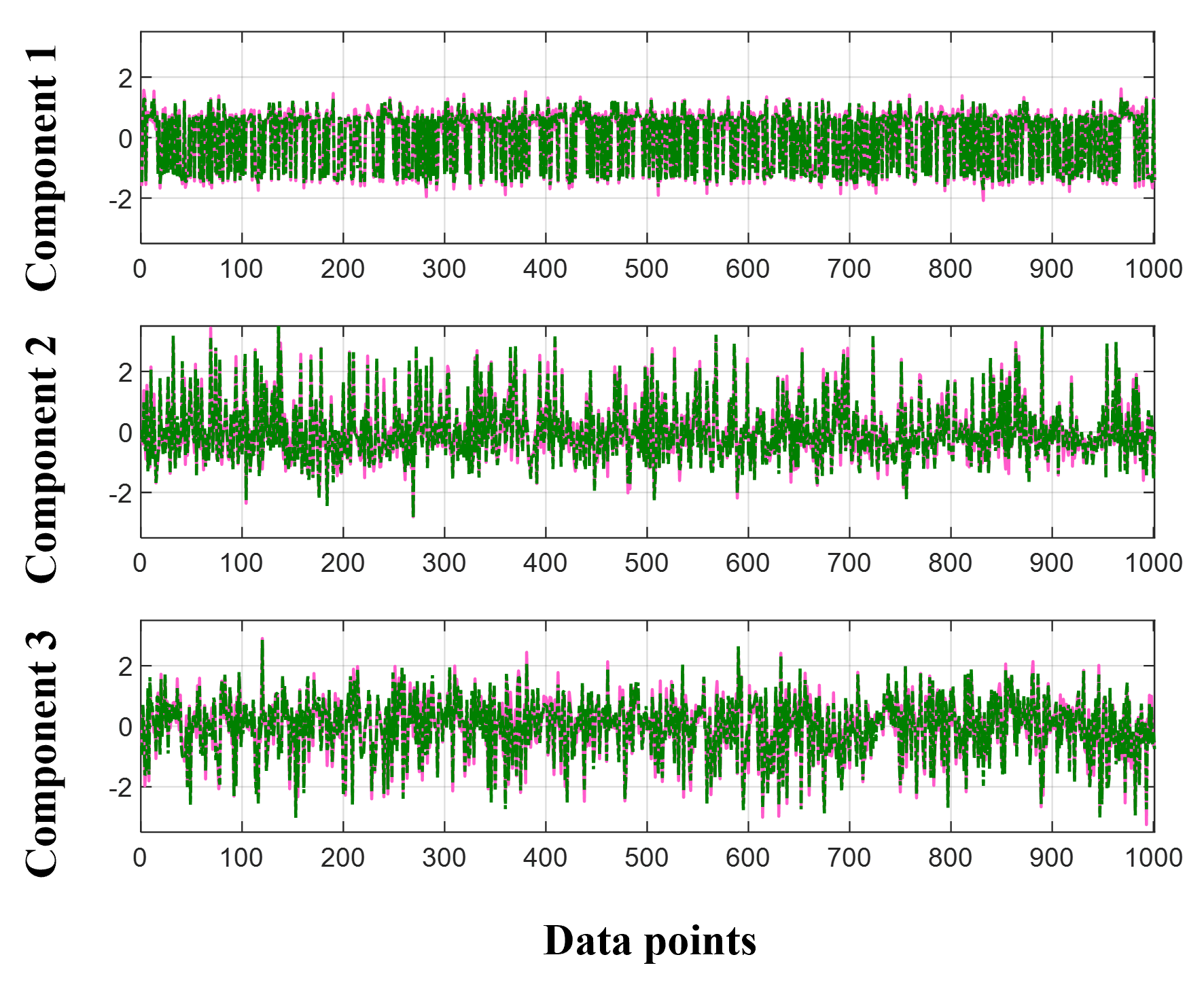}
		\caption{ICA performance of Half-VAE}
		\label{ICA performance of Half-VAE}
	\end{figure}
	
	\begin{figure}[htbp]
		\centering
		\includegraphics[width=0.5\textwidth]{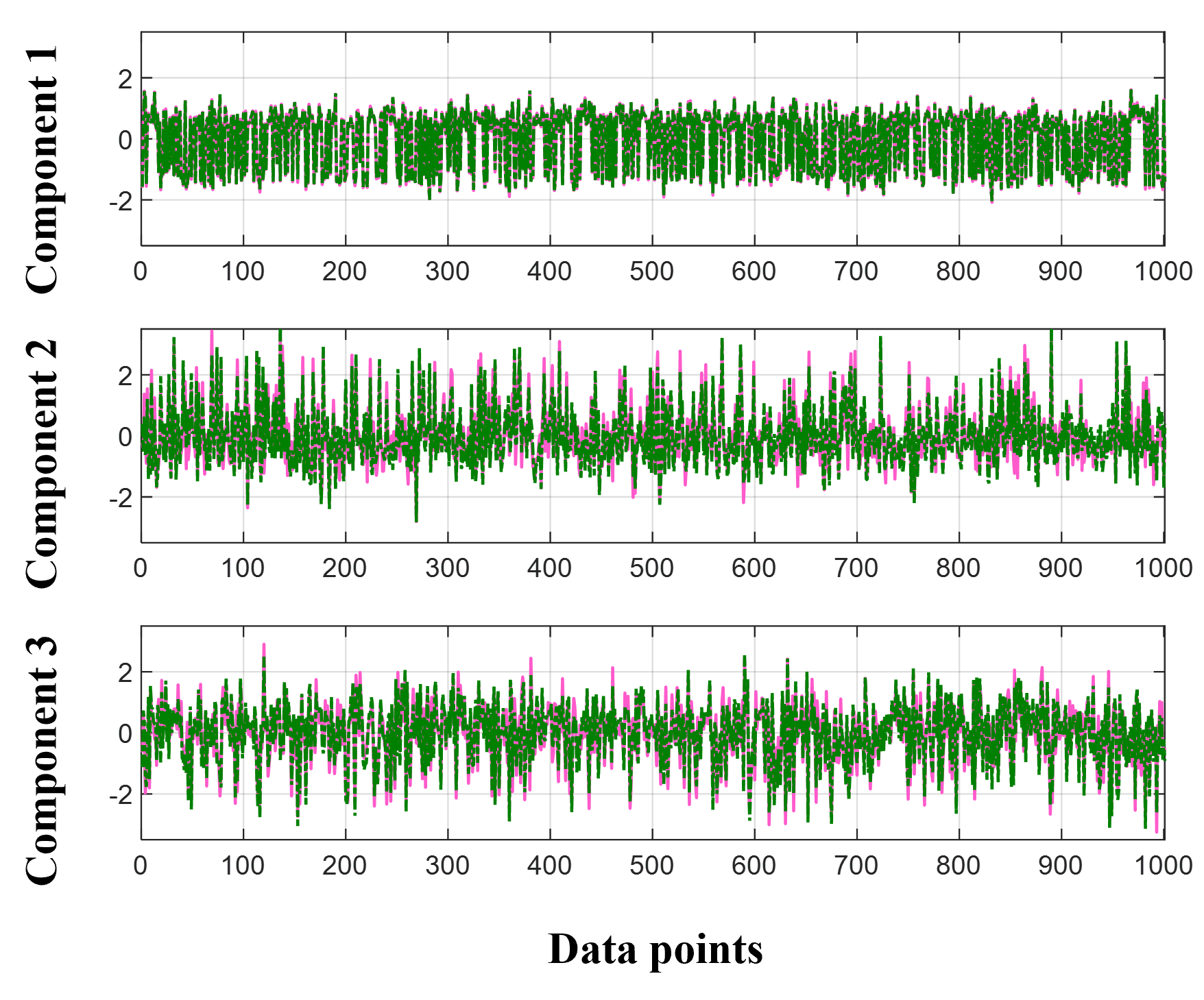}
		\caption{ICA performance of VAE(GMM)}
		\label{ICA performance of VAE)}
	\end{figure}
	
	\begin{figure}[htbp]
		\centering
		\includegraphics[width=0.5\textwidth]{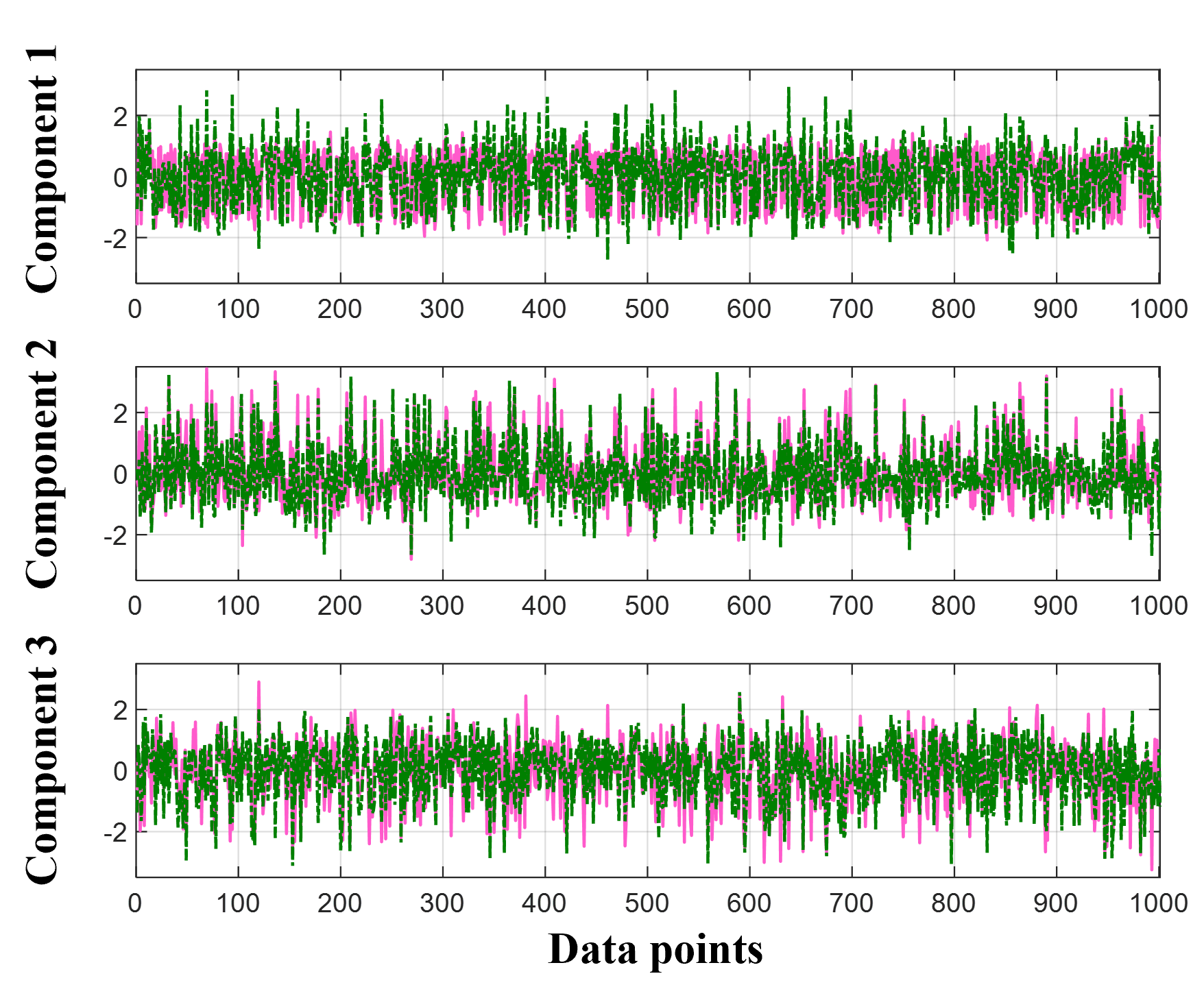}
		\caption{ICA performance of Vanilla VAE}
		\label{ICA performance of Vanilla VAE}
	\end{figure}
	
	\begin{table}[htbp]
		\centering
		\caption{Comparison of the accuracy of inference between Half-VAE, VAE, and Vanilla VAE in solving ICA problems}
		\resizebox{\columnwidth}{!}{ 
			\begin{tabular}{lccc}
				\toprule
				& \textbf{Half-VAE(GMM)} & \textbf{VAE(GMM)} & \textbf{Vanilla VAE} \\
				\midrule
				Component 1 & 0.1174 & 0.0371 & 0.7195 \\
				Component 2 & 0.2023 & 0.3942 & 0.6102 \\
				Component 3 & 0.2342 & 0.3272 & 0.7380 \\
				Mean        & 0.1846 & 0.2528 & 0.6892 \\
				\bottomrule
			\end{tabular}
		}
		\label{tab:comparison_full}
	\end{table}
	
	\section{Discussions and future works}
	This study introduces the Half-VAE, designed to solve ICA problems without explicit inverse mapping. Unlike the traditional VAE architecture, the Half-VAE discards the encoder, with the latent variables directly represented by trainable parameters optimized through the loss function. The research shows that with appropriate prior settings for the latent variables, the Half-VAE is capable of solving ICA problems. In fact, the performance of the Half-VAE in our numerical examples slightly surpasses that of a VAE with an encoder under the same prior conditions.
	
	However, several issues require further investigation. At the current stage of research, the Half-VAE still cannot directly solve ICA problems under underdetermined conditions. Although the Half-VAE avoids explicit inverse mapping, solving underdetermined problems still requires more assumptions and constraints depending on the various conditions (\cite{comon1994independent}; \cite{hyvarinen2001independent}). Additionally, we found that the random initialization of parameters in both the Half-VAE and VAE architectures plays a crucial role in how quickly the models converge. Thus, future work will also investigate the design of effective initialization strategies for both Half-VAE and VAE in solving ICA problems. Finally, beyond the GMM, there are many other suitable priors that can be employed. The selection and comparison of appropriate priors for signals with different characteristics will be addressed and explored in future research.

	
	\section*{Acknowledgments}
	This research was supported by the project Data integration with dialogue systems (1-52JM). We would like to thank the project for its funding and support, which has made this work possible.
	
	\bibliography{ref.bib}

\end{document}